\pdfoutput=1

\documentclass[runningheads]{llncs}
\usepackage{graphicx}
\usepackage{amsmath,amssymb} 
\usepackage{xcolor}
\usepackage{cite}
\usepackage[colorlinks,linkcolor=red,anchorcolor=blue,citecolor=green            ]{hyperref}
\usepackage{booktabs}
\usepackage{array}
\usepackage{tabularx}
\usepackage{bm}

\begin{document}

\title{Identity-Enhanced Network for Facial Expression Recognition} 

\author{Yanwei Li\inst{1,2} \and
Xingang Wang\inst{1} \and
Shilei Zhang\inst{3} \and
Lingxi Xie\inst{4} \and
Wenqi Wu\inst{1,2} \and
Hongyuan Yu\inst{1,2} \and
Zheng Zhu\inst{1,2}}
%

\authorrunning{Li. et al.} 


\institute{Institute of Automation, Chinese Academy of Sciences, Beijing, China \and
University of Chinese Academy of Sciences, Beijing, China \and
IBM Research, Beijing, China \and
The Johns Hopkins University, Baltimore, USA\\
\email{\{liyanwei2017,xingang.wang,wuwenqi2013,yuhongyuan2017,zhuzheng2014\}@ia.ac.cn, slzhang@cn.ibm.com, 198808xc@gmail.com}}

\maketitle

\begin{abstract}
Facial expression recognition is a challenging task, arguably because of large intra-class variations and high inter-class similarities. The core drawback of the existing approaches is the lack of ability to discriminate the changes in appearance caused by emotions and identities. In this paper, we present a novel identity-enhanced network (IDEnNet) to eliminate the negative impact of identity factor and focus on recognizing facial expressions. Spatial fusion combined with self-constrained multi-task learning are adopted to jointly learn the expression representations and identity-related information. We evaluate our approach on three popular datasets, namely Oulu-CASIA, CK+ and MMI. IDEnNet improves the baseline consistently, and achieves the best or comparable state-of-the-art on all three datasets.
\end{abstract}

\section{Introduction}

\begin{figure*}[t]
  \centering
  \includegraphics[width=0.72\linewidth]{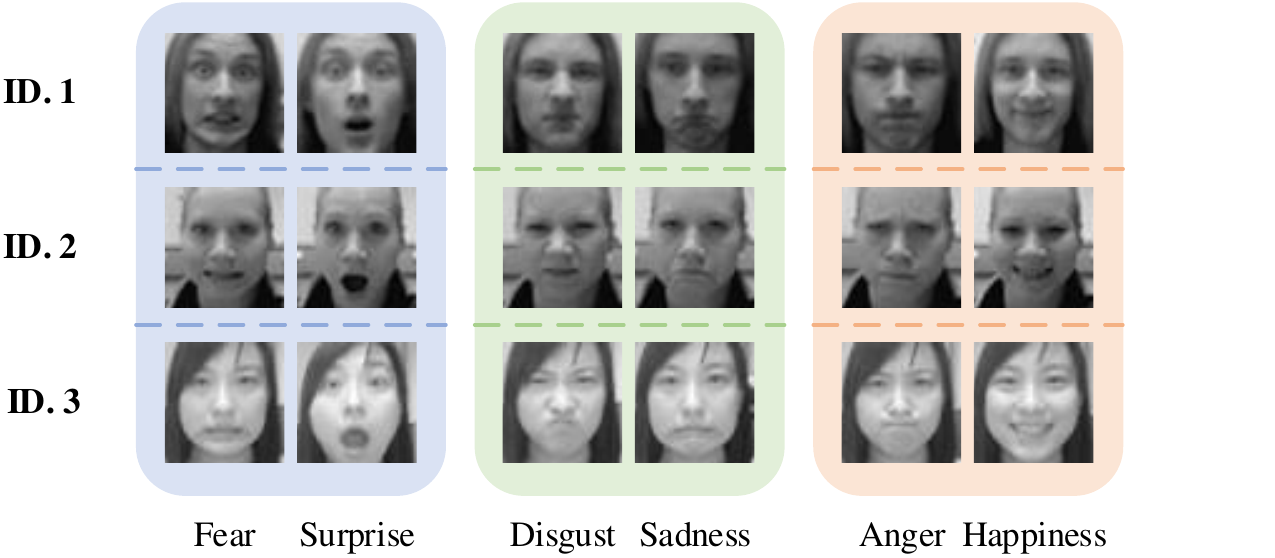}\\
  \caption{Examples of six basic facial expressions, including fear, surprise, disgust, sadness, anger, and happiness. Each row includes examples of six facial expressions with the same identity, and each column includes examples of one specific facial expression with different identities. They are organized by similarities between two facial expressions. Here, we give examples of three people.} \label{fig_intro}
\end{figure*}

Facial expression recognition (FER) is a classic problem in the field of computer vision that attracts a lot of attentions for its wide range of applications in human-computer interaction (HCI) \cite{corneanu2016survey}. FER is challenging mainly due to the large intra-class variations and high inter-class similarities. Identity is a key issue, because the change in identities can bring even heavier variations than the change in expressions. Typical examples are shown in Fig. \ref{fig_intro}. Consequently, this may cause the same expression with different identities to be ranked lower than different expressions. From the perspective of machine learning, this requires projecting the features to another space in which the change in expressions is enhanced and the change in identities is depressed. Some loss functions were designed for this purpose \cite{liu2017adaptive, meng2017identity}, but they were more focused on data-level (reorganizing training data into groups and using metric learning to cancel out ID information) and cannot learn the identities well because of the limited amount of ID cases.

Identity information shares similar facial characteristics with expression features, and these characteristics are concentrated on different facial areas among people. As is presented in Fig. \ref{fig_intro}, the differences of the same expression among people mainly focus on several facial areas (e.g. mouth, nose, eye, and eyebrow), which also reveal their unique identities. However, the relationship between expression features and identity information has seldom been deeply studied in previous works.

For these reasons, we propose a novel identity-enhanced network (IDEnNet) to maximally discriminate identity information from expressions. This is done by jointly learning identity information with expression features and performing expression recognition and identity classification simultaneously. With the enhancement of identity information, the learned expression features are supervised to focus more on several representative facial regions among different expressions and adapt to the changes of identity at the same time. This can be easily learned from Fig. \ref{emosamepeo} and Fig. \ref{emodiffpeo} in ablation study (Sec \ref{ablustu}).

The proposed network structure is illustrated in Fig. \ref{fig_arch}, in which feature extraction groups are adopted to extract identity information together with emotion features from input images and fusion dense block is designed for discriminative expression learning. In order to train this model efficiently, spatial fusion and self-constrained multi-task learning are adopted. Additional loss function whose value decays with the training process is designed to constrain sub-tasks, which facilitates continuous identity learning in the whole training stage.

We introduce multi-level identity information into FER tasks to extend the feature space and enhance ID. The enhancement of identity information also shows effectiveness in well known FER datasets. For example, the accuracy over the baseline is up to \bm{$6.77\%$} on the Oulu-CASIA database \cite{zhao2011facial}.

Overall, the key contributions of this work are summarized as follows:

\begin{itemize}
  \item We propose a novel identity-enhanced network that can effectively discriminate the effect of facial expression changes from that of identity changes with the help of additional identity supervision.
  \item We introduce a loss function whose value decays with learning process to eliminate the over-training risk of auxiliary tasks without early-stopping it during multi-task learning stage.
  \item We present extensive experiments on the three well known FER datasets (namely Oulu-CASIA \cite{zhao2011facial}, CK+ \cite{lucey2010extended}, and MMI \cite{pantic2005web}), and our proposed method achieves better or comparable results of the state-of-the-art methods in all the three datasets.
\end{itemize}

\section{Related Work}
\subsection{Facial Expression Recognition}
Numerous approaches have been proposed to extract emotion features from frames and sequences, such as hand-crafted feature-based method and learned feature-based method \cite{sariyanidi2015automatic,corneanu2016survey}. Traditional hand-crafted features use predesigned appearance \cite{littlewort2011computer, zhao2007dynamic, klaser2008spatio, zhao2011facial}, or geometrical features such as Landmark distances \cite{pantic2006dynamics} to describe relevant facial information. Learned features are usually trained through a joint feature and classification pipeline. Over the past years, deep learning architectures have been developed for end-to-end classification. 

More recently, Jung et al. \cite{jung2015joint} proposed a joint fine-tuning method to fuse geometric features (facial landmarks) with appearance features (images) using deep temporal geometry network and deep temporal appearance network. In \cite{zhao2016peak}, PPDN was proposed to improve the performance of non-peak expressions under the supervision of peak expressions. In \cite{ding2017facenet2expnet}, a regularization function called FaceNet2ExpNet was designed to train expression classification net with the help of pre-trained face net. Related works also considered the impact of identity information in FER tasks. For example, in order to separate identity information from emotion features, Liu et al. \cite{liu2017adaptive} extended triplet loss to (\textit{N+M})-tuplet clusters loss function by incorporating \textit{N} negative sets with the same identity and \textit{M} examples with the same expression. The deep model with (\textit{N+M})-tuplet clusters loss function is subtle, but still hard to train. In contrast with the method mentioned on \cite{liu2017adaptive}, our proposed IDEnNet introduces identity information into FER tasks rather than reducing it. Meng et al. \cite{meng2017identity} tried to alleviate the effect of IDs by adding another FC branch of identity. However, limited amount of ID cases in datasets cannot afford sufficient information for this method.

\subsection{Multi-task Learning}
Multi-task learning has proven its effectiveness in many computer vision problems \cite{zhang2013robust, ranjan2017hyperface}. On one hand, if there is no proper constrain, the performance of main task would be harmed by the over-trained auxiliary tasks. On the other hand, too-early stopping would ``erase'' the features learned by auxiliary tasks. Most of existing multi-task deep models \cite{collobert2008unified} assume similar learning difficulties across all tasks, which are not suitable for our problem. In \cite{ranjan2017hyperface}, many tasks were trained together and balanced by changeless weights in loss function through the whole training stage. In order to appropriately end the learning process of auxiliary tasks, Zhang et al. \cite{zhang2014facial} proposed ``Task-Wise Early Stopping", which terminated auxiliary tasks using highly complex judgements. Such ``early-stopping" operations in our task may erase IDs as training process goes on. Different from previous methods, our proposed loss function (whose value decays with training process) keeps the auxiliary task through the whole training stage in a simple but effective way.

\section{Identity-Enhanced Network}
\begin{figure*}[t]
  \centering
  \includegraphics[width=0.95\linewidth]{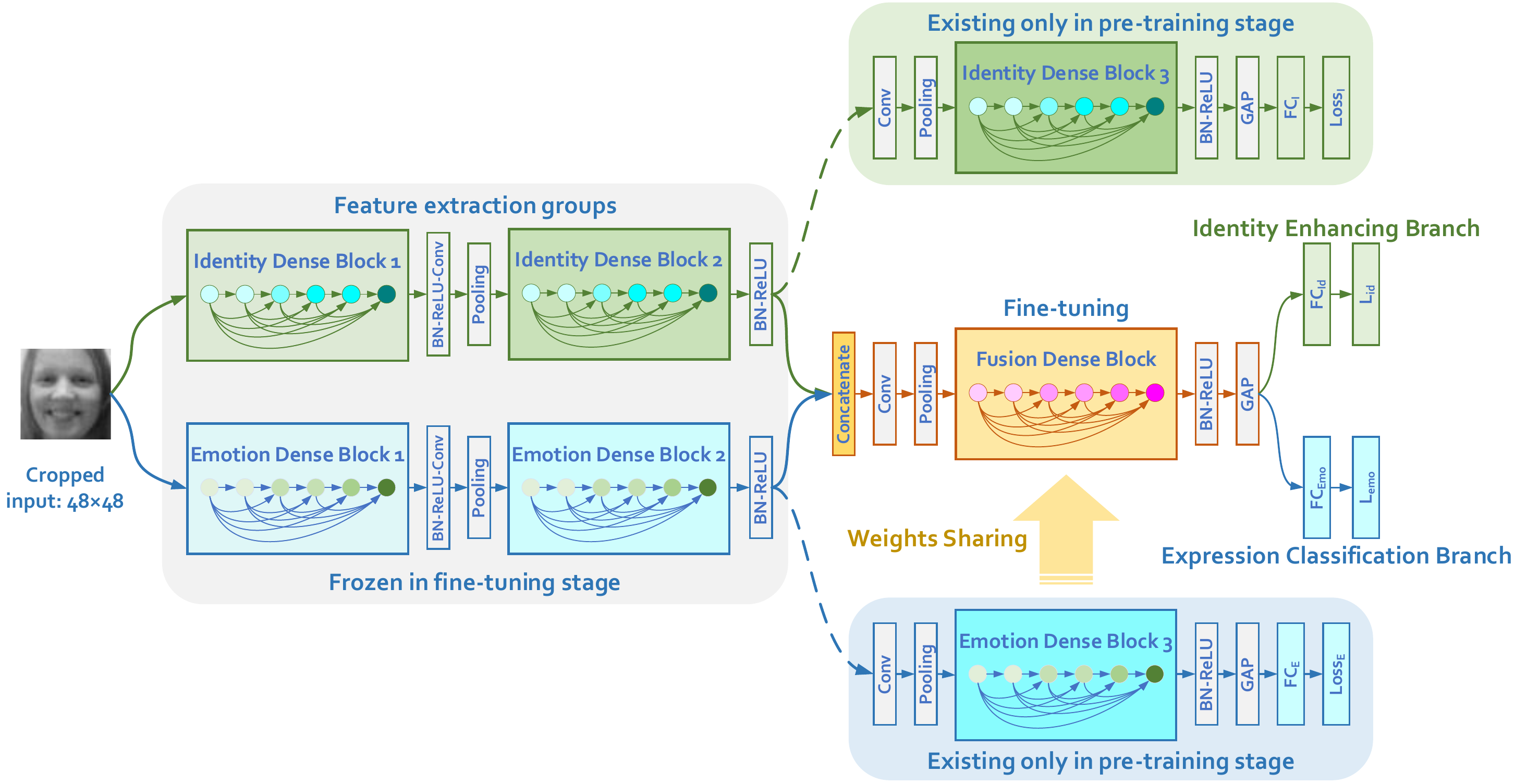}\\
  \caption{The proposed network structure. Here, ``BN" represents batch normalization and ``GAP" means global average pooling. Feature extraction groups (pre-trained) are frozen in the fine-tuning stage to prevent the damage to extracted features during back propagation. Identity Dense Block 3 and Emotion Dense Block 3 are used to train feature extraction groups only in per-training stage. In the testing stage, only the feature extraction groups, fusion dense block (shares weights with pre-trained Emotion Dense Block 3) and expression classification branch are used to recognize a single image.}\label{fig_arch}
\end{figure*}

In this work, we introduce the identity-enhanced network for facial expression recognition. As is illustrated in Fig. \ref{fig_arch}, our designed framework introduces identity information to FER tasks in a fusion way and enhances it by self-constrained multi-task learning (Identity Enhancing Branch in Fig. \ref{fig_arch}). In detail, the proposed ${\mathop{\rm IDEnNet}\nolimits}$ includes five pre-trained DenseNet \cite{8099726} style blocks. Respectively, Identity Dense Block 1 and 2, Emotion Dense Block 1 and 2 are pre-trained with Identity Dense Block 3 and Emotion Dense Block 3 to extract identity features and facial expression features from the input image. And the other dense block (Fusion Dense Block, shares weights with Emotion Dense Block 3) is designed to deeply fuse facial expression representations with identity-related features by several convolution operations.

To clearly illustrate these processes, we start with a cropped image which contains the facial region. It is a $48 \times 48$ grayscale image, then expression feature map ${\mathop{\rm x}\nolimits}_t^{e}$ and identity feature map ${\mathop{\rm x}\nolimits}_t^{i}$ with exactly the same size $H \times W \times D$ are extracted by feature extraction groups after ``Identity Dense Block 2" and ``Emotion Dense Block 2". The following procedures can be decomposed into spatial fusion (Sec \ref{spatialfusion}) and self-constrained multi-task learning section (Sec \ref{wisemtl}).

\subsection{Spatial Fusion}\label{spatialfusion}
Considering the limited number of sequences in FER datasets, we propose a fusion method (``Concatenate", ``Conv", ``Pooling" layers and Fusion Dense Block in Fig. \ref{fig_arch}) to combine facial expression features with identity information. Moreover, two feature streams are combined by convolution, and fusion dense block is designed for a further combination.

The feature map ${{\mathop{\rm y}\nolimits}^{\mathop{\rm cat}\nolimits}}$ after concatenation function can be represented as:
\begin{equation}\label{equ_cat}
 {{\mathop{\rm y}\nolimits}^{\mathop{\rm cat}\nolimits}} = {f^{\mathop{\rm cat}\nolimits}}({\mathop{\rm x}\nolimits}_t^{e},{\mathop{\rm x}\nolimits}_t^{i})
\end{equation}
where ${\mathop{\rm y}\nolimits}^{\mathop{\rm cat}\nolimits} \in {\mathbb{R}^{H \times W \times 2D}}$.

The two feature maps ${\mathop{\rm x}\nolimits}_t^{e}$ and ${\mathop{\rm x}\nolimits}_t^{i}$ are stacked at the same spatial location across the feature channels after concatenation. To better fuse expression features with identity information, we adopt convolution filter $\bm{\mathop{\rm f}\nolimits}$ with size $1 \times 1 \times 2D \times {D^{\mathop{\rm c}\nolimits}}$. Thus, the feature map ${{\mathop{\rm y}\nolimits}^{\mathop{\rm conv}\nolimits}}$ can be written as follows after convolution:
\begin{equation}\label{equ_fea_after_conv}
 {{\mathop{\rm y}\nolimits}^{\mathop{\rm conv}\nolimits}} = {{\mathop{\rm y}\nolimits}^{\mathop{\rm cat}\nolimits}} * \bm{\mathop{\rm f}\nolimits} + b
\end{equation}
where filter ${\bm{\mathop{\rm f}\nolimits} \in {\mathbb{R}^{1 \times 1 \times 2D \times {D^{\mathop{\rm c}\nolimits}}}}}$, bias ${b \in {\mathbb{R}^{1 \times 1 \times 2D}}}$, ${D^{\mathop{\rm c}\nolimits}}$ is the number of output channels. Here, the filter $\bm{\mathop{\rm f}\nolimits} $ is adopted to reduce the dimensionality and combine two feature maps ${\mathop{\rm x}\nolimits}_t^{e},{\mathop{\rm x}\nolimits}_t^{i}$ at the same spatial location.

The advantage of convolutional fusion is that the filter $\bm{\mathop{\rm f}\nolimits}$ and bias $b$ here can be optimized to combine expression features with identity information in a subtle way after back propagation. And the concatenated features are further learned by the following ``Fusion Dense Block", which enhances several representative facial regions between similar expressions and encourages the network to be more adaptive to identity change. The effect of spatial fusion is evaluated  and visualized in the ablation study (Sec \ref{ablustu}).

\subsection{Self-Constrained Multi-task Learning}\label{wisemtl}
The above method introduces identity information stream to the network in a fusion way, but it may not use up the capability of two streams. Consequently, we propose a self-constrained multi-task learning method (Identity Enhancing Branch in Fig. \ref{fig_arch}) to enhance the identity characteristics contained in expression features, which boosts the network performance in FER tasks.

Firstly, the proposed network architecture is reused, as shown in Fig. \ref{fig_arch}. Next, we add another linear fully connected layer (${\mathop{\rm FC}\nolimits}_{\mathop{\rm Id}\nolimits}$) to the network for identity classification, which is located below the global average pooling function. Note that the identity classification is an auxiliary task utilized to enhance identity information. Thus, an additional function is needed to constrain the identity classification task before it is over-trained and harms the FER task.

Here we propose a novel method called self-constrained multi-task learning to constrain the auxiliary task. This method uses additional loss function ($L_{{\mathop{\rm id}\nolimits}}$ in Fig. \ref{fig_arch}) whose value decays quickly with training process to prevent over-training of the auxiliary task. In the designed function, the loss of expression recognition, the main task, is cross-entropy loss, which can be written as:
\begin{equation}\label{equ_emo_loss}
  {L_{{\mathop{\rm emo}\nolimits} }} =  - \sum\limits_{i = 1}^c {{y_i}\log \left( {{p_i}} \right)}
\end{equation}
where ${{y_i}}$ is $i$-th value (0 or 1) of the ground truth label, and ${{p_i}}$ is the $i$-th value of softmax activation function.

To constrain identity classification training process, we introduce the loss function of identity ${L_{{\mathop{\rm id}\nolimits}}}$ by extending binary Focal loss \cite{Lin_2017_ICCV} to multi-class style:
\begin{equation}\label{equ_id_loss}
 {L_{{\mathop{\rm id}\nolimits}}} =  - \sum\limits_{j = 1}^c {\alpha {{(1 - {p_j})}^\gamma{y_j}}\log ({p_j})}
\end{equation}
where ${{y_j}}$ is $j$-th value (0 or 1) of the ground truth label, ${{p_j}}$ is the $j$-th value of softmax activation function, $\alpha$ and $\gamma$ are hyper parameters used to constrain training process, with $\alpha$ and $\gamma \ge 0$.

Obviously, the format of ${L_{{\mathop{\rm id}\nolimits}}}$ is similar with ${L_{{\mathop{\rm emo}\nolimits}}}$, which just multiply a constrain parameter ${\alpha {{(1 - {p_j})}}^\gamma }$. With the training process going deeper, the value of ${\alpha {{(1 - {p_j})}}^\gamma }$ decays quickly due to the increase of $p_j$. Consequently, the training process of identity classification is constrained by this parameter. Thus, the joint loss function can be defined as:
\begin{equation}\label{equ_emo_id_loss}
 {L_{{\mathop{\rm joint}\nolimits}}} = {L_{{\mathop{\rm emo}\nolimits}}} + {L_{{\mathop{\rm id}\nolimits}}}
\end{equation}

Theoretically and experimentally we find that due to the use of parameter ${\alpha {{(1 - {p_j})}}^\gamma }$, the auxiliary task (identity classification) is constrained in the whole training stage. In the designed loss, auxiliary tasks are considered as background roles, which means loss value of identity classification could be much smaller than the main task, but bigger than zero. With the use of designed loss function, auxiliary tasks can be utilized to supervise the training process as well as enhance robustness of the main task without stopping in the whole training stage. In practice, we find that the network performs best when $\gamma  = 15$ and $\alpha  = 0.1$ (we elaborate it in supplementary material). Furthermore, self-constrained multi-task learning can be extended to multiple auxiliary tasks learning by adding other loss functions in a similar form to ${L_{{\mathop{\rm id}\nolimits}}}$.

\subsection{Network Architecture}\label{netarch}
Each dense block in the network contains six layers which consists of BN-ReLU-Conv($1\times1$)-BN-ReLU-Conv($3\times3$), and concatenates input with the output of each layer. Here, each convolution layer includes 12 filters. All of the pooling layers except global average pooling in the network are $2\times2$ average pooling with stride 2. The dimension of ${{\mathop{\rm FC}\nolimits}_{\mathop{\rm Id}\nolimits}}$ and ${{\mathop{\rm FC}\nolimits}_{\mathop{\rm Emo}\nolimits}}$ adjust to the number of identities and facial expressions respectively. Depending on whether to use spatial fusion and self-constrained multi-task learning, the proposed method can be divided into four sub-networks: Original Network (baseline without additional identity information, contains Emotion Dense Block 1 and 2, Fusion Dense Block, and Expression Classification Branch), ${\mathop{\rm IDEnNet}\nolimits}_I$ (contains Original Network and additional Identity Enhancing Branch), ${\mathop{\rm IDEnNet}\nolimits}_F$ (contains Feature extraction groups, Fusion Dense Block, and Expression Classification Branch), and ${\mathop{\rm IDEnNet}\nolimits}_{IF}$ (contains all of the dense blocks and branches). 

\section{Experiments}
We empirically demonstrate the effectiveness of proposed method on three public databases: Oulu-CASIA \cite{zhao2011facial}, CK+ \cite{lucey2010extended}, and MMI database \cite{pantic2005web}. The following shows details and results of these experiments as well as ablation study (Sec \ref{ablustu}).

\subsection{Dataset Description}\label{datadesc}
\subsubsection{Oulu-CASIA.}
The Oulu-CASIA database \cite{zhao2011facial} includes 480 sequences consist of 80 subjects and six basic expressions (anger, disgust, fear, happiness, sadness, and surprise) taken under dark, strong and weak illumination conditions. The same way as before \cite{ding2017facenet2expnet}, we use videos with strong condition captured by a VIS camera in this experiment. Each sequence of the database begins with neutral and ends with a peak expression. Thus, only the last three frames of each sequence are kept for training and testing. We adopt 10-fold cross validation protocol for training and validating.
\subsubsection{CK+.}
CK+ \cite{lucey2010extended} is a widely used database for FER tasks, which includes 118 subjects consist of seven emotions: anger, contempt, disgust, fear, happiness, sadness, and surprise. There are 327 image sequences in the database, which begin with neutral and end with a peak expression. Similar to the Oulu-CASIA database, we only use last three frames of each sequence and adopt 10-fold cross validation protocol for training and validating.
\subsubsection{MMI.}
For further experiments, we use MMI database \cite{pantic2005web}, which contains 312 image sequences from 30 individuals with six basic expressions (same expressions with Oulu-CASIA). Here, we select 205 sequences captured in a front view. Different from Oulu-CASIA and CK+, each sequence in MMI starts from a neutral face, reaches the peak in the middle and returns to neutral expression at the end. So we use three frames in the middle of each sequence following the protocol of 10-fold cross validation. Moreover, this database includes individuals who wear glasses or pose expression non-uniformly. Consequently, the facial expression recognition task is relatively challenging compared to other databases.

\subsection{Preprocessing}\label{preproce}
To reduce the variation in face scale, we apply SeetaFace face detection \cite{wu2017funnel} and alignment \cite{zhang2014coarse} modules for face detection and landmark detection. Then, the region of interest based on the coordinate of detected landmarks is cropped and resized to $60 \times 60$. Due to the limited images in FER datasets, several data augmentation procedures are employed to alleviate the over-fitting problem. Following \cite{jung2015joint}, we horizontally flip the whole cropped images at first. Then, the region of interest in each image is rotated by each angle in $\{ - 15^\circ , - 10^\circ , - 5^\circ ,5^\circ ,10^\circ ,15^\circ \}$. Thus, we obtain a new dataset 14 times larger than the original one: original, flipped, rotated with six angles, and their flipped version.

\subsection{Implementation Details}\label{impdetail}
For network training, we firstly pre-trained two networks respectively with exactly the same architecture (as is illustrated in Fig. \ref{fig_arch}) in FER+ dataset \cite{barsoum2016training} and CASIA-WebFace dataset \cite{yi2014learning} for facial expression and identity recognition (our accuracy is up to $81.22\%$ on CASIA-WebFace with no bells and whistles). Both of them are trained with initial learning rate set to 0.1, which is divided by 10 at $50\%$ and $75\%$ of the total training epochs.

In the fine-tuning stage, we only fine-tune the layers below concatenation operation, and freeze the feature extraction groups. In this stage, we randomly crop images to $48 \times 48$ when training. A single center crop with size $48 \times 48$ is used for testing. We optimize the network using Stochastic Gradient Descent with a momentum of 0.9. The initial learning rate for fine-tuning is 0.01, and decreased by 0.1 at $50\%$ and $75\%$ of the total training epochs. The mini-batch size, weight decay parameter and drop out rate are set to 128, 0.0001 and 0.5.

As described in Sec \ref{netarch}, there are original network and other three types of proposed ${\mathop{\rm IDEnNet}\nolimits}$. Here, we adopt original network as our baseline, and do comparative experiments based on it.

\subsection{Results}\label{dataresults}
Among all the compared databases, our proposed IDEnNet outperforms the state-of-the-art methods including handcraft-based methods (LBP-TOP \cite{zhao2007dynamic}, and HOG 3D \cite{klaser2008spatio}), video-based methods (MSR \cite{ptucha2011manifold}, AdaLBP \cite{zhao2011facial}, Atlases \cite{guo2012dynamic}, STM-ExpLet \cite{liu2014learning}, and DTAGN \cite{jung2015joint}), and CNN-based methods (3D-CNN \cite{liu2014deeply}, 3D-CNN-DAP\cite{liu2014deeply}, DTAGN \cite{jung2015joint}, PPDN\cite{zhao2016peak}, GCNet \cite{kim2017deep}, and FN2EN \cite{ding2017facenet2expnet}).

\subsubsection{Oulu-CASIA.}
Table \ref{accoulu} reports the average accuracy of 10-fold cross validation results. Compared with other state-of-the-art algorithms, our proposed algorithm achieves substantial improvement. As is shown, our method is superior to the previous best performance achieved by FN2EN \cite{ding2017facenet2expnet}, with a gain of $6.24\%$. Respectively, ${\mathop{\rm IDEnNet}\nolimits}_I$, ${\mathop{\rm IDEnNet}\nolimits}_F$, and ${\mathop{\rm IDEnNet}\nolimits}_{IF}$ have improved $0.48\%$, $4.21\%$ and $6.77\%$ over the baseline of $87.18\%$. We can find that ${\mathop{\rm IDEnNet}\nolimits}_I$ have slight improvement over the original one, while the network with spatial fusion improves a lot. What's more, the result shows that self-constrained multi-task learning after spatial fusion further learns difference between similar facial expressions. This is consistent with the analysis in ablation study (Sec \ref{ablustu}).

\begin{table}[t]
\centering
\caption{Average accuracy of Oulu-CASIA database. }\label{accoulu}
\newcolumntype{Y}{>{\centering\arraybackslash}X}
\begin{tabularx}{0.65\linewidth}{ Y || c }
 \hline
    Method & Average Accuracy (\%) \\
 \hline\hline
    HOG 3D\cite{klaser2008spatio} & $70.63$\\
    AdaLBP\cite{zhao2011facial} & $73.54$\\
    STM-ExpLet\cite{liu2014learning} & $74.59$\\
    Atlases\cite{guo2012dynamic} & $75.52$\\
    DTAGN\cite{jung2015joint} & $81.46$\\
    PPDN\cite{zhao2016peak} & $84.59$\\
    GCNet\cite{kim2017deep} & $86.39$\\
    FN2EN\cite{ding2017facenet2expnet} & $87.71$\\
 \hline
    Original Network(baseline) & $87.18$\\
    \textbf{Ours(\bm{${\mathop{\rm IDEnNet}\nolimits}_I$})} & \bm{$87.66$}\\
    \textbf{Ours(\bm{${\mathop{\rm IDEnNet}\nolimits}_F$})} & \bm{$91.39$}\\
    \textbf{Ours(\bm{${\mathop{\rm IDEnNet}\nolimits}_{IF}$})} & \textcolor{red}{\bm{{$93.95$}}}\\

 \hline
\end{tabularx}
\end{table}


\begin{table}[t]
\centering
\caption{Average accuracy of CK+ database, 7 expressions. }\label{accck}
\newcolumntype{Y}{>{\centering\arraybackslash}X}
\begin{tabularx}{0.65\linewidth}{ Y || c }
 \hline
    Method & Average Accuracy (\%)\\
 \hline\hline
    3D-CNN\cite{liu2014deeply} & $85.9$\\
    LBP-TOP\cite{zhao2007dynamic} & $88.99$\\
    MSR\cite{ptucha2011manifold} & $91.4$\\
    HOG 3D\cite{klaser2008spatio} & $91.44$\\
    3D-CNN-DAP\cite{liu2014deeply} & $92.4$\\
    STM-ExpLet\cite{liu2014learning} & $94.19$\\
    IACNN\cite{meng2017identity} & $95.37$\\
    (\textit{N+M}) Softmax\cite{liu2017adaptive} & $97.1$\\
    DTAGN\cite{jung2015joint} & $97.25$\\
    GCNet\cite{kim2017deep} & $97.93$\\
 \hline
    Original Network(baseline) & $98.23$\\
    \textbf{Ours(\bm{${\mathop{\rm IDEnNet}\nolimits}_I$})} & \textcolor{red}{\bm{$98.46$}}\\
    \textbf{Ours(\bm{${\mathop{\rm IDEnNet}\nolimits}_F$})} & \bm{$98.42$}\\
    \textbf{Ours(\bm{${\mathop{\rm IDEnNet}\nolimits}_{IF}$})} & \bm{$98.42$}\\

 \hline
\end{tabularx}
\end{table}

\subsubsection{CK+.}
 In Table \ref{accck}, we compare our method with other state-of-the-art algorithms which use 7 expressions for training and validating. Our proposed method shows a better performance than all other compared algorithms including GCNet\cite{kim2017deep} which compares given images with generative faces to generate contrastive representation for classification. Due to the high accuracy of benchmark algorithm, the improvement of our proposed method in CK+ is less obvious than in Oulu-CASIA. As the number of individuals in CK+ is bigger, the contribution of multi-task learning (${\mathop{\rm IDEnNet}\nolimits}_I$) is relatively greater than that in Oulu-CASIA when compared with spatial fusion method (${\mathop{\rm IDEnNet}\nolimits}_F$), which could also be learned from Table \ref{accck}. The slight drop ($0.04\%$) of ${\mathop{\rm IDEnNet}\nolimits}_{IF}$ is mainly due to some random factors, {\em e.g.} network initialization. This can be verified by using weaker networks on the CK+ dataset, as is reported in Table \ref{Tab:CK_Accuracy}.


\begin{table}[t]
\centering
\caption{Average accuracy of MMI database. }\label{accmmi}
\newcolumntype{Y}{>{\centering\arraybackslash}X}
\begin{tabularx}{0.65\linewidth}{ Y || c }
 \hline
    Method & Average Accuracy (\%)\\
 \hline\hline
    3D-CNN\cite{liu2014deeply} & $53.2$\\
    LBP-TOP\cite{zhao2007dynamic} & $59.51$\\
    HOG 3D\cite{klaser2008spatio} & $60.89$\\
    3D-CNN-DAP\cite{liu2014deeply} & $63.4$\\
    DTAGN\cite{jung2015joint} & $70.24$\\
    IACNN\cite{meng2017identity} & $71.55$\\
    CSPL\cite{zhong2012learning} & $73.53$\\
    STM-ExpLet\cite{liu2014learning} & $75.12$\\
    (\textit{N+M}) Softmax\cite{liu2017adaptive} & $78.53$\\
    GCNet\cite{kim2017deep} & $81.53$\\
 \hline
    Original Network(baseline) & $73.62$\\
    \textbf{Ours(\bm{${\mathop{\rm IDEnNet}\nolimits}_I$})} & \bm{$73.82$}\\
    \textbf{Ours(\bm{${\mathop{\rm IDEnNet}\nolimits}_F$})} & \bm{$79.41$}\\
    \textbf{Ours(\bm{${\mathop{\rm IDEnNet}\nolimits}_{IF}$})} & \textcolor{red}{\bm{$80.19$}}\\

 \hline
\end{tabularx}
\end{table}

\subsubsection{MMI.}
 Due to the small size of this database, it is difficult for deep networks to learn features. Thus, we adopt 16 layers instead of 40 layers as backbone network. As is presented in Table \ref{accmmi}, the previous top accuracy was only $81.53\%$.  In more detail, ${\mathop{\rm IDEnNet}\nolimits}_{IF}$ improves $6.57\%$ over the baseline, while ${\mathop{\rm IDEnNet}\nolimits}_I$ and ${\mathop{\rm IDEnNet}\nolimits}_F$ increase by $0.20\%$ and $5.79\%$ respectively. Here, we strongly recommend using large database in practice. Totally, even under the limitation of this database, the proposed network still has a better ability for FER.

\subsection{Ablation Study}\label{ablustu}
In this subsection, we will reveal the effect of backbone network as well as the amount of IDs in training data. Moreover, in order to better demonstrate the effectiveness of our proposed network, heatmaps after ``Fusion Dense Block"  of expressions with same identity and expressions with different identities in Oulu-CASIA database \cite{zhao2011facial} are visualized.

\subsubsection{Backbone Network.}
In order to prove our generalization performance on different backbone network and  databases, we apply different backbones (16-, 22- and 40-layer DenseNets) on CK+  \cite{lucey2010extended} and FER+ \cite{barsoum2016training} dataset. For the lack of identity label of images in FER+ dataset, we only adopt ${\mathop{\rm IDEnNet}\nolimits}_F$ for experiments. As is elaborated in  Table \ref{Tab:CK_Accuracy} and \ref{Tab:FER_Accuracy}, our proposed IDEnNet consistently boosts the performance on two datasets whatever the backbone network is. What's more, IDEnNet also improves $2.85\%$ on FER+ without data augmentation.

\begin{table*}[htb]
\centering
\caption{ Average accuracy of CK+ in different backbones,.}
\label{Tab:CK_Accuracy}
\begin{tabular}{l||c|c|c}
\hline
Method & $16$ layers & $22$ layers & $40$ layers  \\
\hline\hline
Original & $91.19\%$ & $95.16\%$ & $98.23\%$ \\
\hline
\bm{${\mathop{\rm IDEnNet}\nolimits}_I$} & \bm{$92.13\%$} & \bm{$96.57\%$} &  \textcolor{red}{\bm{$98.46\%$}}  \\
\hline
\bm{${\mathop{\rm IDEnNet}\nolimits}_F$} & \bm{$92.87\%$} & \bm{$96.49\%$} & \bm{$98.42\%$} \\
\hline
\bm{${\mathop{\rm IDEnNet}\nolimits}_{IF}$} & \textcolor{red}{\bm{{$93.46\%$}}} & \textcolor{red}{\bm{{$96.63\%$}}} & \bm{{$98.42\%$}}  \\
\hline
\end{tabular}
\end{table*}

\begin{table*}[h!]
\centering
\caption{ Average accuracy of FER+ in different backbones.}
\label{Tab:FER_Accuracy}
\begin{tabular}{l||c|c|c}
\hline
Method & $16$ layers & $22$ layers & $40$ layers  \\
\hline\hline
Original & $76.37\%$ & $79.93\%$ & $80.11\%$ \\
\hline
\bm{${\mathop{\rm IDEnNet}\nolimits}_F$} & \textcolor{red}{\bm{{$78.42\%$}}} & \textcolor{red}{\bm{{$80.67\%$}}} & \textcolor{red}{\bm{{$82.96\%$}}} \\
\hline
\end{tabular}
\end{table*}

\subsubsection{Identity Contribution.}
\begin{figure}[h]
  \centering
  \includegraphics[width=0.55\linewidth]{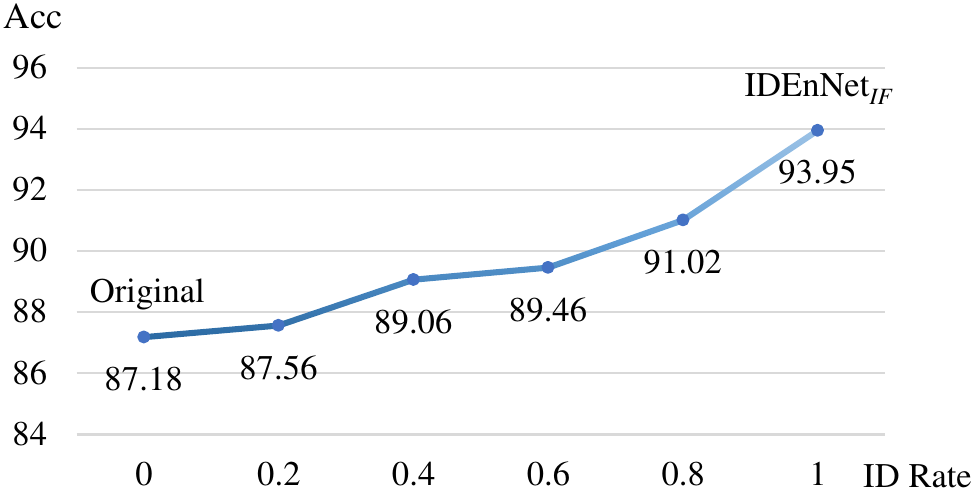}\\
  \caption{Accuracy of IDEnNet on Oulu-CASIA with different ID rates.}\label{ID_rate}
\end{figure}

To evaluate the contribution of identity, We perform an ablation study on Oulu-CASIA by using all training images but only a subset of ID information. Unsurprisingly, more IDs are utilized, better results are obtained, as is presented in Fig. \ref{ID_rate}. The curve in Fig. \ref{ID_rate} also proves the effectiveness of ID information when adopting IDEnNet.

\subsubsection{Expressions with Same Identity.}
\begin{figure}[t]
  \centering
  \includegraphics[width=0.75\linewidth]{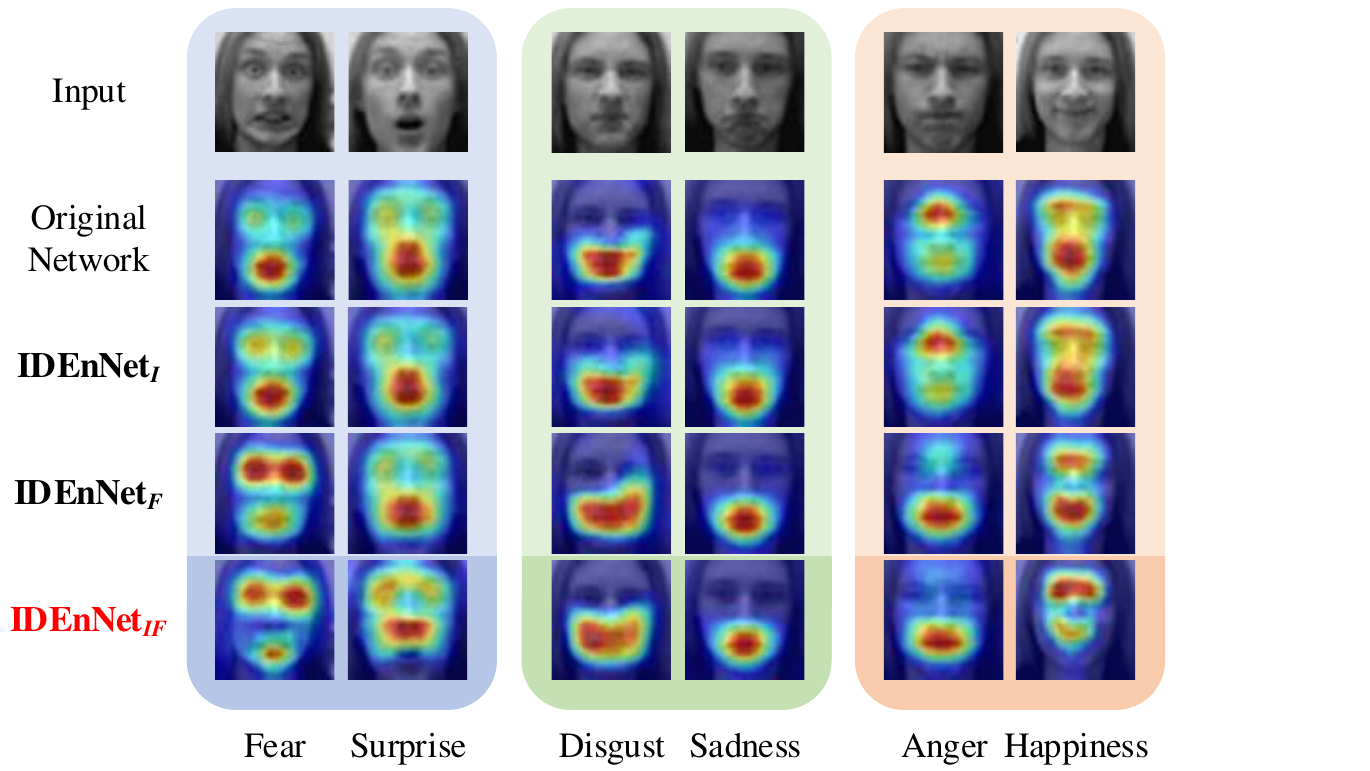}\\
  \caption{Heatmaps of facial expressions with same identity in Oulu-CASIA database\cite{zhao2011facial}. The images in the first row are used as input. The second to fifth rows are heatmaps under original network, ${\mathop{\rm IDEnNet}\nolimits}_I$, ${\mathop{\rm IDEnNet}\nolimits}_F$, and ${\mathop{\rm IDEnNet}\nolimits}_{IF}$ respectively. Blue and red in each heatmap represent the low and high response value. They are organized in pairs according to the similarity between two expressions as well as corresponding regions in the heatmaps under the original network.} \label{emosamepeo}
\end{figure}

To further explore the effect of the proposed network for expressions in the same identity, we draw the heatmaps in each proposed method. As is presented in Fig. \ref{emosamepeo}, heatmaps in the ``Original Network" show different areas which are highly respond to the corresponding expressions. These heatmaps are drawn under the original network which means identity information in the network has not been enhanced. It can be seen that the areas with high response are mainly concentrated on the mouth, nose, eyes and eyebrows, which is consistent with our prior knowledge.

With the addition of identity information, the high response areas trend to be more focused and change their corresponding priorities according to different expressions, as is illustrated in Fig. \ref{emosamepeo}. From two similar expressions (Fear and Surprise), we can find that our proposed ${\mathop{\rm IDEnNet}\nolimits}_I$, ${\mathop{\rm IDEnNet}\nolimits}_F$, and ${\mathop{\rm IDEnNet}\nolimits}_{IF}$ adjust their attention to eyes and mouth when recognizing fear, while concentrating more on nose wing and eyebrows when identifying surprise. And this also can be learned from other pairs of similar classes. Moreover, the addition of identity information encourages the network to use features which represent the specific identity. For different expressions, the identity characteristics tend to locate in different facial areas, which also encourages the network to use different facial regions. Thus, our proposed ${\mathop{\rm IDEnNet}\nolimits}$ have the ability to focus more on distinguishable facial regions when classifying similar expression classes. That is to say, ${\mathop{\rm IDEnNet}\nolimits}$ enlarge the inter-class difference of facial expressions due to the enhancement of identity information.

\subsubsection{Expressions with Different Identities.}
As well known, the performance of different people have great difference when doing the same facial expression. Here, we compare performance of the proposed ${\mathop{\rm IDEnNet}\nolimits}$ for these people.

\begin{figure}[t]
  \centering
  \includegraphics[width=0.7\linewidth]{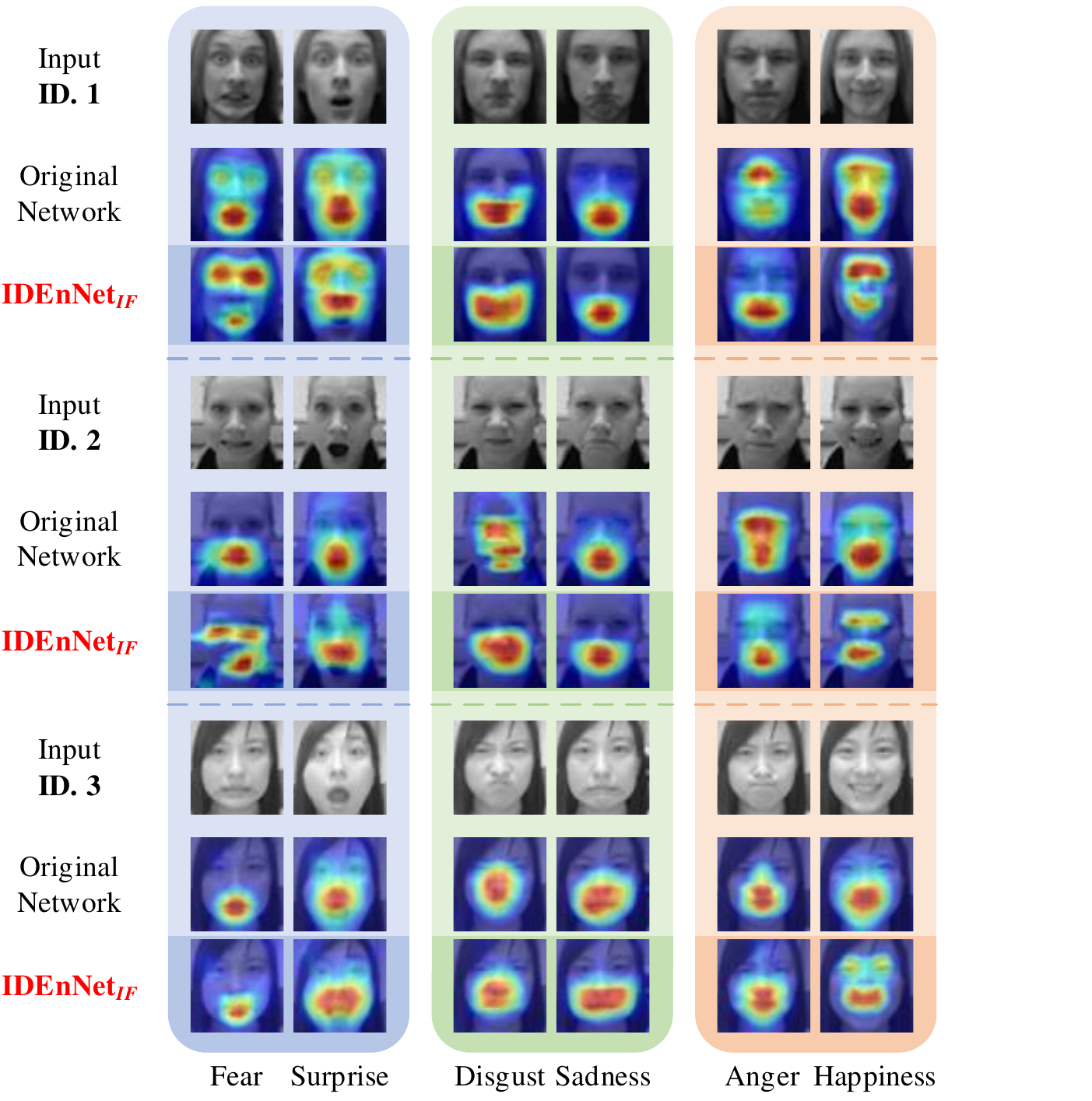}\\
  \caption{Heatmaps of facial expressions with different identities in Oulu-CASIA database \cite{zhao2011facial}. The images in the first, fourth, and seventh row are input images of three people. The following two rows after each input are heatmaps under original network and ${\mathop{\rm IDEnNet}\nolimits}_{IF}$ respectively. Blue and red in each heatmap represent the low and high response value. They are organized in pairs according to the similarities between two expressions as well as corresponding regions in the heatmaps under the original network.}\label{emodiffpeo}
\end{figure}

\begin{figure}[t]
  \centering
  \includegraphics[width=0.6\linewidth]{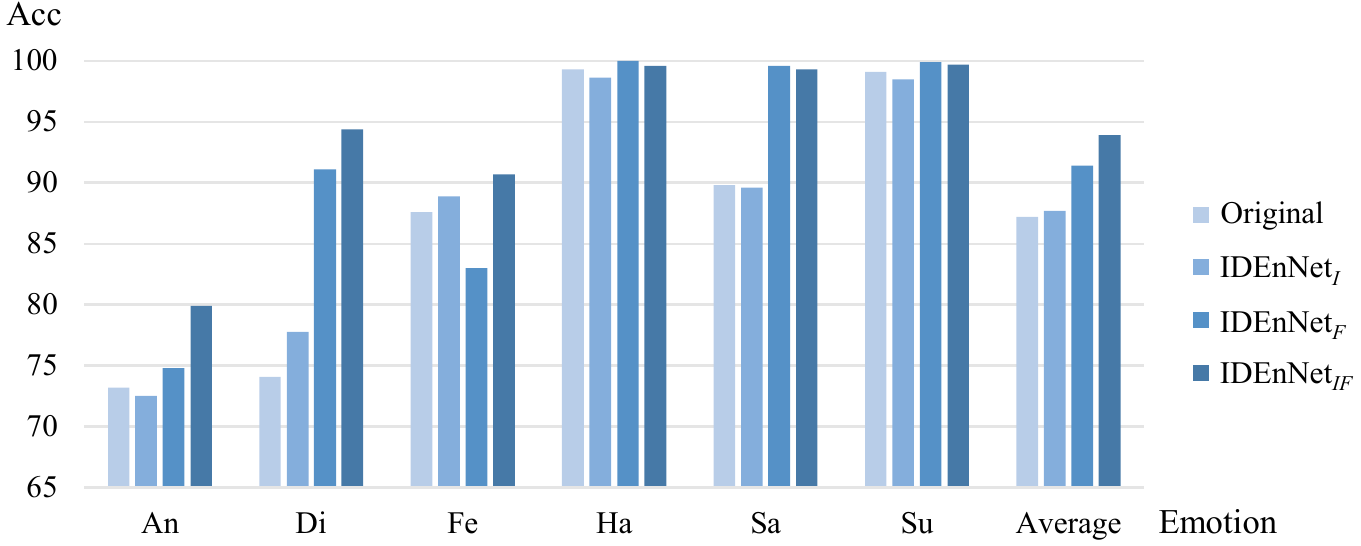}\\
  \caption{Comparison of accuracy in the Oulu-CASIA database \cite{zhao2011facial} according to each emotion among the proposed networks.}\label{ouluresultemo}
\end{figure}

Concerning the same expression of three people, we can learn from Fig. \ref{emodiffpeo} that ${\mathop{\rm IDEnNet}\nolimits}_{IF}$ trends to use similar facial regions of people for the same expression, adapting slightly to different identities at the same time. For example, original network mainly uses mouth, nose wing and eyebrows areas of the ID. 1 man to recognize his happiness, but mainly uses mouth and nose wing areas of both the ID. 2 and ID. 3 woman to represent the same expression. With the addition of identity information, our proposed ${\mathop{\rm IDEnNet}\nolimits}_{IF}$ mainly concentrate on both mouth and eyes areas of three people to recognize their happiness. The proposed ${\mathop{\rm IDEnNet}\nolimits}_{IF}$ also shows the same effect on other expressions. As for classes of fear and surprise, ${\mathop{\rm IDEnNet}\nolimits}_{IF}$ inclines to focus on mouth areas for fear and nose wing areas for the other. Furthermore, the proposed network also uses the eyes areas for fear expression of ID. 1 man and ID. 2 woman, and uses the eyes with wider areas to recognize the surprise expression of the ID. 1 man, which can be attributed to adaptation of ${\mathop{\rm IDEnNet}\nolimits}_{IF}$ to different identities. With the enhancement of identity information, our proposed ${\mathop{\rm IDEnNet}\nolimits}$ trends to learn the key representation which is helpful to the identity recognition, and the representative features are often located in similar regions for different people. This is also confirmed by Fig. \ref{ouluresultemo}, which shows that ${\mathop{\rm IDEnNet}\nolimits}_{IF}$ has a great improvement over the original network, especially in the cases of anger, disgust and sadness. Therefore, we can draw the conclusion that the proposed ${\mathop{\rm IDEnNet}\nolimits}$ trends to focus more on several similar facial regions of different people for the same expression. Namely, it reduce the intra-class variations of the same expression among people.

\begin{figure}[t]
  \centering
  \includegraphics[width=0.65\linewidth]{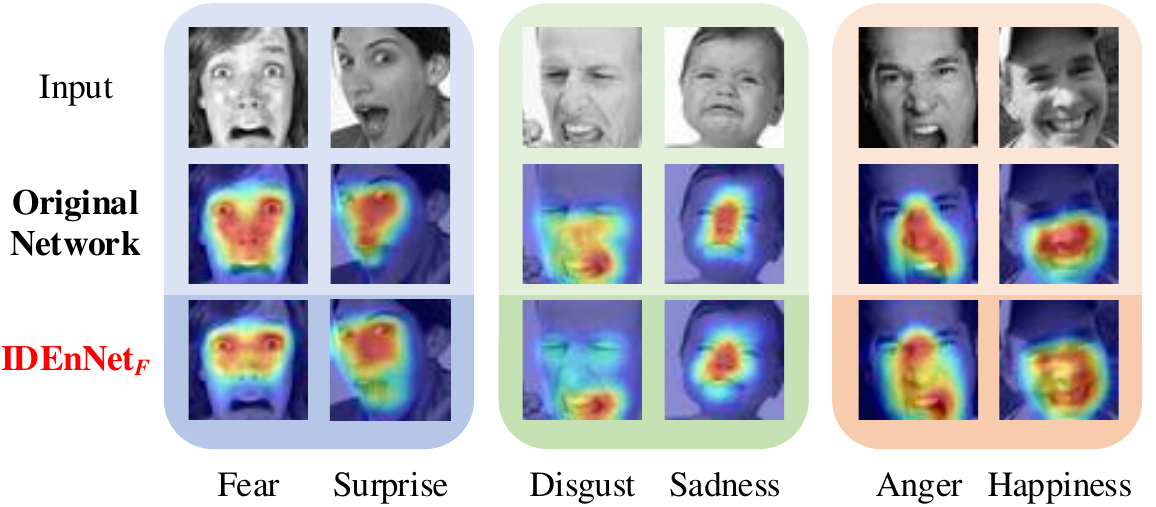}\\
  \caption{Heatmaps of facial expressions with weakly-aligned faces in FER+ database \cite{barsoum2016training}.}\label{FERresultemo}
\end{figure}

We also visualize some profile face images from FER+. As shown in Fig. \ref{FERresultemo}, IDEnNet works well in dealing with these weakly-aligned cases. In addition, the attention areas in Fig. \ref{emodiffpeo} and \ref{FERresultemo} share the same trend, indicating that IDEnNet learns to focus on semantic regions even in profile faces rather than learns a face template.

\section{Conclusions}
In this paper, we propose a novel identity-enhanced network which explicitly learns from identity information so that the network can focus on representative facial regions between similar expressions as well as adapt to different identities. In the proposed IDEnNet, feature extraction groups are adopted to extract identity information together with expression features and fusion dense block is designed for identity-related expression learning. In order to train this network efficiently, spatial fusion and self-constrained multi-task learning are adopted, which also enables the identity cues to be combined in expression recognition. Extensive experiments on several datasets demonstrate the effectiveness of our proposed method.

Furthermore, the feature enhancement approach proposed in this work has potential application value for other  fine-grained recognition tasks. The recognition performance could be much better if the pose information is utilized for supervision in a feature enhancement method when recognizing different kinds of birds or dogs.

\section*{Acknowledgement}
This work has been supported by the National Key Research and Development Program of China No. 2018YFD0400902 and National Natural Science Foundation of China under Grant 61573349.

\clearpage
%
%
%
\bibliographystyle{splncs04}
\bibliography{accv_cr}
\end{document}